\documentclass{article}

\usepackage{PRIMEarxiv}

\usepackage[utf8]{inputenc} 
\usepackage[T1]{fontenc}    
\usepackage{hyperref}       
\usepackage{url}            
\usepackage{booktabs}       
\usepackage{amsfonts}       
\usepackage{nicefrac}       
\usepackage{microtype}      
\usepackage{lipsum}
\usepackage{fancyhdr}       
\usepackage{graphicx}       
\graphicspath{{media/}}     

\usepackage{amsmath}
\usepackage{graphicx}
\usepackage{subcaption}

\title{Tokenize Features, Enhancing Tables: The FT-TabPFN Model for tabular Classification
}

\author{
  Liuquangao\\
  Shenyang Institute of Automation \\
  Chinese Academy of Sciences \\
  Shenyang\\
  \texttt{liuquangao@sia.cn}\\
   \And
  Yangwei \\
  Shenyang Institute of Automatio \\
  Chinese Academy of Sciences \\
  Shenyang\\
  \texttt{epicard@163.com} \\
  \AND
  Liangchen \\
  Shenyang Institute of Automation \\
  Chinese Academy of Sciences \\
  Shenyang \\
  \texttt{liangchen@sia.cn} \\
  \And
  Panglonglong \\
  Shenyang Institute of Automation \\
  Chinese Academy of Sciences \\
  Shenyang \\
  \texttt{panglonglong@sia.cn} \\
  \And
  Zouzhuozhang \\
  Shenyang Institute of Automation \\
  Chinese Academy of Sciences \\
  Shenyang \\
  \texttt{zouzhuozhang@sia.cn} \\
}

\begin{document}
\maketitle

\begin{abstract}
Traditional methods for tabular classification usually rely on supervised learning from scratch, which requires extensive training data to determine model parameters. However, a novel approach called Prior-Data Fitted Networks (TabPFN) has changed this paradigm. TabPFN uses a 12-layer transformer trained on large synthetic datasets to learn universal tabular representations. This method enables fast and accurate predictions on new tasks with a single forward pass and no need for additional training. Although TabPFN has been successful on small datasets, it generally shows weaker performance when dealing with categorical features. To overcome this limitation, we propose FT-TabPFN, which is an enhanced version of TabPFN that includes a novel Feature Tokenization layer to better handle classification features. By fine-tuning it for downstream tasks, FT-TabPFN not only expands the functionality of the original model but also significantly improves its applicability and accuracy in tabular classification. Our full source code is available for community use and development (see \href{https://github.com/liuquangao/FT-TabPFN}{link}).
\end{abstract}


\section{Introduction}
Tabular data , structured in rows and columns, where rows denote samples and columns denote features, is prevalent in fields like business analytics\cite{xia2014review}\cite{seng2010intelligent}, finance\cite{ngai2011application,carmona2022no,sattarov2023findiff}, and healthcare\cite{di2023explainable,menegotto2021computer}. 

For a long time, Gradient Boosted Decision Trees (GBDT)\cite{friedman2001greedy}have dominated tabular classification due to their short training time and strong robustness. Recently, a transformative model known as TabPFN\cite{hollmann2023tabpfn} has redefined this landscape. 
The model is trained on datasets generated based on various predefined priors to facilitate fast and accurate classification. Moreover, it approximates probabilistic inference for the priors in a single forward pass. In the new classification task, TabPFN does not require parameter optimization or model fitting on downstream training data. Instead, it takes both the training and test sets as inputs and derives predictions for the test set through contextual interactions. A recent large-scale tabular classification study\cite{mcelfresh2024neural} showed that TabPFN achieves average state-of-the-art classification on small datasets (number of samples less than 2000). However, TabPFN may have limited application due to design constraints, particularly when dealing with categorical features. TabPFN is reported to perform exceptionally well on small tabular datasets that have purely numerical features, but generally shows weaker performance when dealing with categorical features\cite{hollmann2023tabpfn,mcelfresh2024neural}. In this study, we focus on improving the performance of TabPFN on categorical features. Overall, our contributions are as follows: 

\textbf{Contribution 1.} We propose FT-TabPFN, a refined version of TabPFN, which includes a novel Feature Tokenization layer. This layer is designed to enhance the processing of categorical features within tabular data, enabling more effective handling of diversity in data types.

\textbf{Contribution 2.} We further introduce a regularization mechanism for feature identifiers within the Feature Tokenization layer. This regularisation helps to maintain independence and uniqueness between features , thus improving the performance and robustness of the model.

\textbf{Contribution 3.} We apply FT-TabPFN through fine-tuning on downstream tasks and verify its effectiveness through experiments.

\section{Related Work}
\label{sec:relatedwork}
\subsection{Tabular Classification}
\subsubsection{traditional models}
In traditional machine learning, logistic regression\cite{pregibon1981logistic}, support vector machine (SVM)\cite{hearst1998support} and decision tree\cite{quinlan2014c4} are widely used because of their simplicity and efficiency. Logistic regression is suitable for binary classification problems and has a simple, easy-to-interpret model, but it is difficult to deal with non-linear relationships.SVMs effectively deal with non-linear problems through kernel tricks, but are computationally inefficient on large-scale datasets. Decision trees are easy to understand and implement, but are prone to overfitting, especially on complex datasets with many features.
To overcome these limitations, integrated learning methods such as Random Forests\cite{breiman2001random} and Gradient Boosting Machines (GBM)\cite{friedman2001greedy} have been introduced. Random forest reduces overfitting and improves classification performance by voting or averaging multiple decision trees.GBMs, especially XGBoost\cite{chen2016xgboost}, LightGBM\cite{ke2017lightgbm}, and CatBoost\cite{prokhorenkova2018catboost}, have become the dominant method for processing tabular data by improving model accuracy through stepwise optimisation.
\subsubsection{Deep learning Models}

Deep learning has had notable success in areas such as image, audio and text processing, and is increasingly being applied to tabular data. Recent innovations include models such as SAINT \cite{somepalli2021saint}, FT-Transformer \cite{gorishniy2021revisiting}, NPT \cite{kossen2021self}, t2g-former \cite{yan2023t2g}, TabCaps\cite{chen2022tabcaps}, and TabNet \cite{arik2021tabnet}, which are trained from scratch and designed to effectively capture complex patterns in data.

\subsection{TabPFN}
\begin{figure}[t!]
    \centering
    \includegraphics[width=1\linewidth]{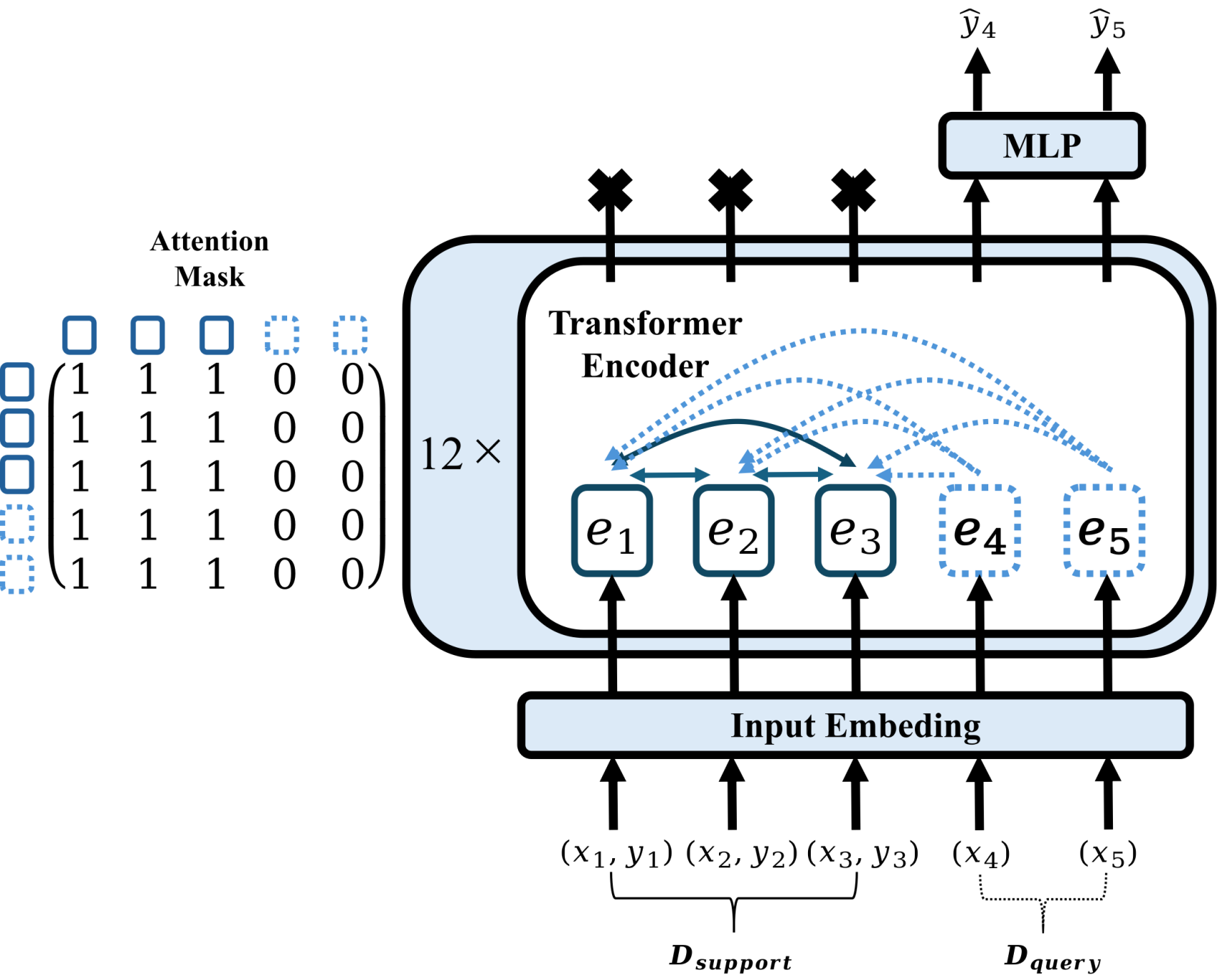}
    \caption{\textbf{ A brief visualisation of TabPFN.
}There are three support samples and two query samples. The x and y of the support samples are embedded together in a vector. The x of the query samples are embedded in a vector. The y of the query samples are waiting for prediction.}
    \label{figTabPFN}
\end{figure}

\begin{figure}[t!]
    \centering
    \includegraphics[width=0.8\linewidth]{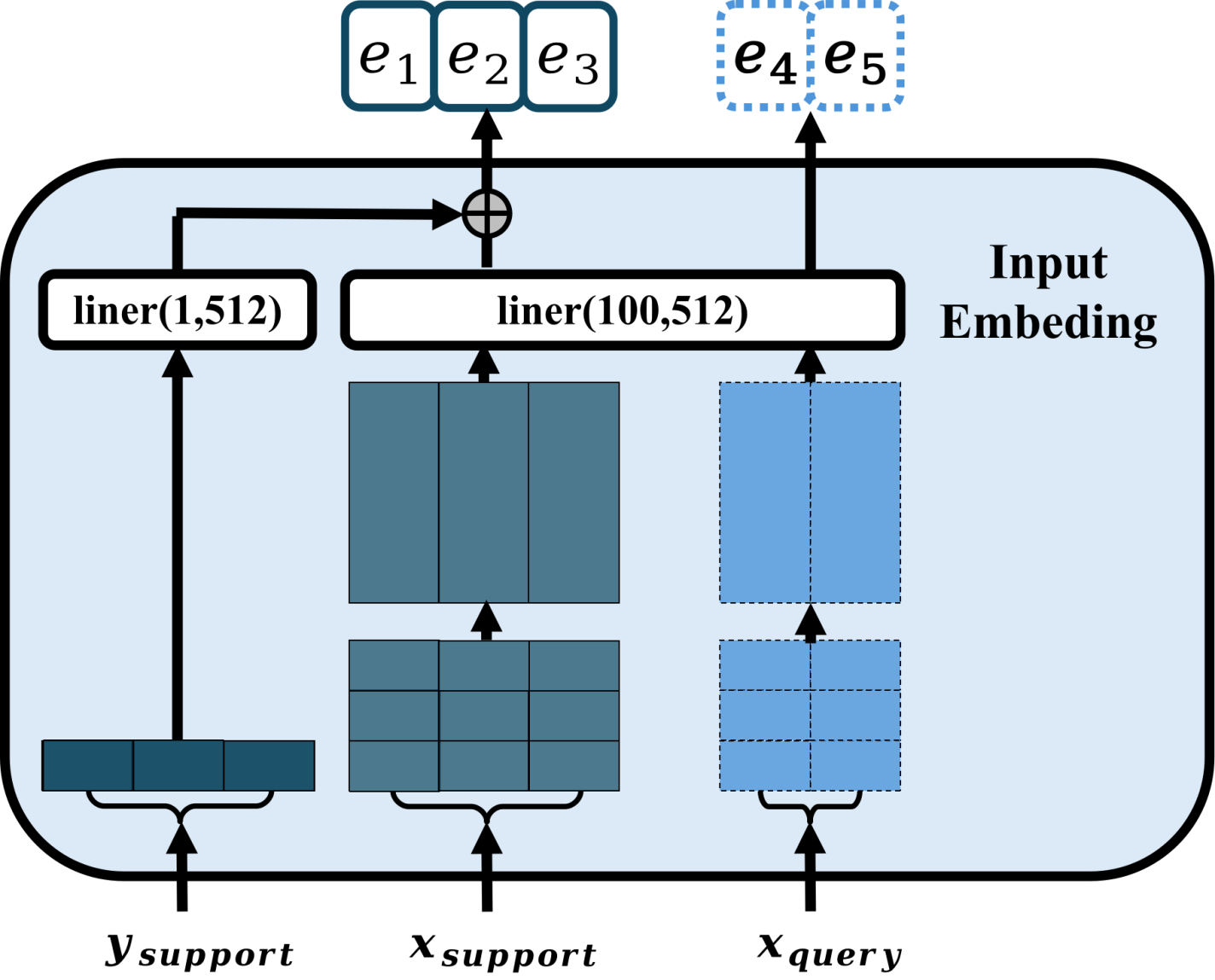}
    \caption{\textbf{ Input Embeding module of TabPFN.} The feature vector x of query samples is expanded from its original dimensions to 100 dimensions through zero-padding, and then mapped to a 512-dimensional space via a linear layer liner(100,512) to form the query sample embedding. Support samples undergo a similar process for their x. However, the y of support samples is mapped to the same 512-dimensional space independently through another linear layer liner(1,512). Finally, the embeddings of x and y for support samples are added together to form the final support sample embedding.}
    \label{figinputembeding}
\end{figure}

The backbone of the TabPFN\cite{hollmann2023tabpfn} is a stack of 12-layer transformer encoders as shown in the Fig.\ref{figTabPFN} . TabPFN is pre-trained on a large manually curated dataset for tabular data. Unlike traditional models that are trained from scratch for each new task, TabPFN processes both labelled samples (referred to as 'support samples') and unlabelled samples (referred to as 'query samples') at once, using in-context learning methods similar to \cite{von2023transformers,li2023transformers,garg2022can}.
That is, TabPFN takes the entire data set as input and makes predictions on the unlabelled samples in it with a single forward pass. This approach allows for immediate and effective classification without the need for re-training the model on new datasets.

\subsection{Tokenization}
In Natural Language Processing (NLP), Tokenization is a fundamental technique for segmenting text into easily processable units of information called tokens. These units - usually words, phrases or characters - form the basis for subsequent processing and model training. The core purpose of Tokenization is to deconstruct complex textual data into a form that models can understand and process, enabling them to capture the structure and semantics of language. This technique is widely used in a variety of NLP tasks such as machine translation\cite{vaswani2017attention}, sentiment analysis\cite{medhat2014sentiment}, text classification\cite{devlin2018bert} and question and answer systems\cite{shao2019transformer,zhao2020condition}. By converting text into token sequences, NLP models can perform language understanding and generation tasks more easily, thus improving processing efficiency and performance. In this work, we extend the concept of tokenization to tabular data, aiming to enhance model processing efficiency and performance by adapting this technique to handle diverse data formats systematically.This is described in detail in Section \ref{sec:method}.

\section{Motivation}

Tabular data are inherently heterogeneous, characterised by a mixture of different columns from different domains, each with unique data types and distributions\cite{borisov2022deep,villaizan2024graph,li2024tree}. This heterogeneity presents unique challenges in data processing, particularly in the differential handling required for categorical and continuous features. Unlike continuous features, which correlate with mathematical metrics and directly support operations such as arithmetic and statistical analysis, categorical features represent discrete categories or groups with no intrinsic numerical relationships, requiring special encoding techniques.

The limitations of TabPFN in distinguishing between these feature types motivated our research. As shown in Fig.\ref{figinputembeding} TabPFN does not distinguish between categorical and continuous features during the embedding process, using a uniform linear transformation for all feature types. This approach can lead to suboptimal representations for categorical features, as it imposes a false ordinal relationship where none exists.

Inspired by Natural Language Processing (NLP), where tokenization breaks down text into manageable and semantically rich components ("tokens"), we propose a novel feature processing methodology for tabular Feature Tokenization. Our method transforms each feature into a high-dimensional, computationally feasible format similar to word embeddings by treating each feature as a "word" and each data sample as a "sentence". The aim of this approach is to improve the representational power of the features so that the model can learn and understand patterns in the data more efficiently.
\section{Method}
\label{sec:method}

\begin{figure*}[t!]
    \centering
    \includegraphics[width=0.8\textwidth]{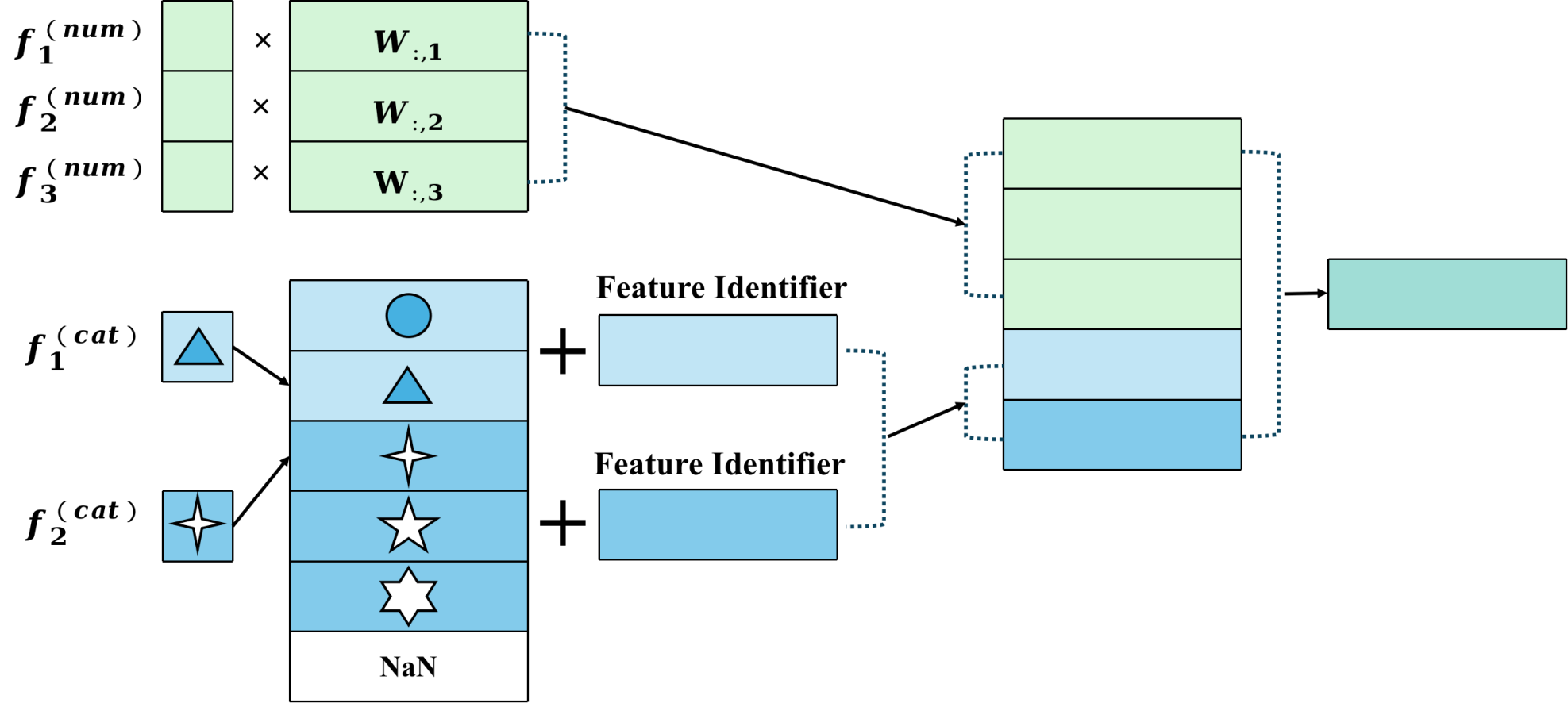}
    \caption{\textbf{Feature Tokenization Layer.} This innovative layer processes continuous and categorical features distinctly, transforming them into feature tokens which are then aggregated to construct the final sample embedding.}
    \label{figFT}
\end{figure*}

In order to address the limitations of the traditional TabPFN approach to the handling of categorical features, we have developed FT-TabPFN. This extended model includes a novel feature tokenization layer that replaces the standard linear transformation layer $\textit{liner}(100,512)$ used in the original TabPFN input embedding module (see Fig. \ref{figinputembeding}). This modification allows for more nuanced and effective differentiation and processing of categorical and continuous features.

Consider a feature-target pair $(x,y)$, where $x= \left ( x^{(\textit{num})},x^{(\textit{cat})}\right ) \in \mathbb{X}$ represents numerical  $x^{(\textit{num})}$  and categorical  $x^{(\textit{cat})}$  features.  Specifically, $x^{(\textit{num})}=\{f_i^{(\textit{num})}\}_{i=1}^n$ consists of $n$ numerical features and $x^{(\textit{cat})}=\{f_j^{(\textit{cat})}\}_{j=1}^m$ consists of $m$ categorical features. In the Feature Tokenization layer (Fig. \ref{figFT}), each feature—numerical ($x^{(\textit{num})}$) or categorical ($x^{(\textit{cat})}$)—is converted into a set of feature tokens, $T \in \mathbb{R}^{(m+n) \times d}$, where $m$ and $n$ represent the number of categorical and numerical features, respectively, and $d=512$ denotes the dimensionality of the embedding space.

For numerical features:
\begin{equation}
T_i^{(\textit{num})} = f_i^{(\textit{num})} \cdot W_{i}^{(\textit{num})},
\end{equation}
where $W^{(\textit{num})} \in \mathbb{R}^{m \times d}$ is a weight matrix initially derived from the parameters of the original linear layer $\textit{liner}(100,512)$. This initialization is designed to retain the robust performance of TabPFN on numerical features. To prevent overfitting, this weight matrix is kept frozen during the training process.

For categorical features:
\begin{equation}
T_j^{(\textit{cat})} = W^{(\textit{cat})}[g(f_j^{(\textit{cat})})] + I_j,
\end{equation}
where $W^{(\textit{cat})} \in \mathbb{R}^{(N+1) \times d}$ is a token lookup table accommodating $N$ categories plus an additional token for NaN values (defined as zeros). The function $g$ maps each feature value to its corresponding index in $W^{(\textit{cat})}$. The feature identifiers $I \in \mathbb{R}^{m \times d}$, akin to positional encodings in NLP, enhance the model’s capacity to manage feature-specific characteristics.

In order to further improve the ability of FT-TabPFN to distinguish different categorical features, we performed orthogonal regularisation of the feature identifiers. The mathematical expression of the regularisation process is as follows:
\begin{equation}
\text{OrthogonalLoss} = \sum_{i \neq j} G_{ij}^2,
\end{equation}
where \(G = I^T I\) is the Gram matrix of the feature identifiers after normalizing each identifier to have unit norm, \(G_{ij}\) represents the dot product between the \(i^{th}\) and \(j^{th}\) feature identifiers (thus measuring their cosine similarity), and the sum is taken over all pairs of different identifiers (i.e., \(i \neq j\)).

Our approach is conceptually consistent with FT-Transformer \cite{gorishniy2021revisiting}, which also converts all features into tokens. But there are actually obvious differences: FT-Transformer lacks the processing of NaN, and uses the attention mechanism on them to classify samples after turning the features into tokens. Our model has NaN processing. After the features are turned into tokens, they are directly added to obtain the sample embedding for subsequent processing. Furthermore, unlike FT-Transformer, which includes all types of feature biases, our model selectively applies feature identifiers only to categorical features.


\section{Experiments}
\begin{table}[h]
\caption{\textbf{Dataset details and training settings}}
\label{table:datasets_details}
\centering
\begin{tabular}
{@{\hskip 0mm}
c@{\hskip 1mm}
c@{\hskip 1mm}
c@{\hskip 1mm}
c@{\hskip 1mm}
c@{\hskip 1mm}
c@{\hskip 1mm}
c@{\hskip 1mm}
c@{\hskip 1mm}
c@{\hskip 0mm}}
\toprule
Dataset Name    &   Samples&    Numerical&    Categorical&  Class&  Epoch& Learning rate\\
\midrule
cmc             &   1473&                2&             8&       3&    30&  1e-3\\
credit-approval &    690&                6&             10&      2&    30&  1e-3\\
tic-tac-toe     &    958&                0&             10&      2&    30&  1e-2\\
cylinder-bands  &    540&               18&             22&      2&    30& 1e-2\\
dresses-sales	&    500&                1&             12&      2&    30&  1e-3\\
\bottomrule
\end{tabular}
\end{table}

\subsection{Datasets And Models}

\begin{table}[h]
\centering
\caption{\textbf{ROC AUC OVO results on the six datasets} We only conducted experiments on different versions of FT-TabPFN, and the results of other models are from \cite{hollmann2023tabpfn}.}
\label{my_tabular_results_table}
\resizebox{\textwidth}{!}{
  \begin{tabular}
{@{\hskip 0mm}
l@{\hskip 1mm}
l@{\hskip 1mm}
l@{\hskip 1mm}
l@{\hskip 1mm}
l@{\hskip 1mm}
l@{\hskip 1mm}
l@{\hskip 1mm}
l@{\hskip 1mm}
l@{\hskip 1mm}
l@{\hskip 1mm}
l@{\hskip 0mm}}
\toprule
{} & LightGBM & CatBoost & XGBoost & ASKL2.0 & AutoGluon & TabPFN$_{n.e.}$ & TabPFN & FT-TabPFN$_{n.r.}$ &FT-TabPFN$_{n.i.}$ & FT-TabPFN \\
\midrule
cmc                  & 0.7288 & 0.7256 & 0.7299 & \textbf{0.7378} & 0.7331 & 0.7233 & 0.7274 & 0.7218 & 0.7276 & 0.7314 \\
credit-approval      & 0.9415 & 0.9389 & 0.9422 & 0.9406 & 0.9415 & 0.9253 & 0.9322 & 0.9358  & 0.9261 & \textbf{0.9425} \\
tic-tac-toe          & 0.9988 & 0.9992 & \textbf{1} & 0.9943 & \textbf{1} & 0.9547 & 0.9759 & 0.9920  & 0.8975& \textbf{1} \\
cylinder-bands       & 0.8556 & 0.8757 & 0.8782 & 0.8718 & \textbf{0.8878} & 0.8314 & 0.8336 & 0.8392  & 0.8508 & 0.8530 \\
dresses-sales        & 0.5593 & 0.5696 & 0.5823 & 0.5705 & 0.5507 & 0.5333 & 0.5376 & 0.5841  & 0.5810 &\textbf{0.6169} \\
\midrule
Wins AUC OVO & 0 & 0 & 1 & 1 & 2 & 0 & 0 & 0 & 0 & \textbf{3} \\
M. rank AUC OVO & 5.1 & 5.4 & \textbf{2.6} & 4.2 & 3.3 & 9.6 & 8.0 & 6.2 & 8.0 & \textbf{2.6} \\
Mean AUC OVO & 0.8168 & 0.8218 & 0.8265 & 0.8230 & 0.8226 & 0.7936 & 0.8014 & 0.8157 &0.7954 & \textbf{0.8288} \\
\bottomrule
\end{tabular}
}
\end{table}

We evaluated our FT-TabPFN on five distinct datasets: cmc, credit-approvals, tic-tac-toe, cylinder-tape, and dress-sales, all of which contain categorical features. Details of these data and the training settings are shown in Table.\ref{table:datasets_details}. Previous studies have shown that standard TabPFN models underperform on these datasets due to their categorical features\cite{hollmann2023tabpfn}.

\textbf{Models Overview:}
\begin{itemize}
    \item \textbf{FT-TabPFN:} The complete model with feature identifiers and orthogonal regularization.
    \item \textbf{FT-TabPFN${n.i.}$:} Without feature identifiers.
    \item \textbf{FT-TabPFN${n.r.}$:} With feature identifiers but without orthogonal regularization.
    \item \textbf{TabPFN$_{n.e.}$:} This model configuration denotes the outcomes of a standard TabPFN following a single forward computation pass. It serves as a baseline within our analysis.
    \item \textbf{TabPFN:} This version of TabPFN utilizes an ensemble approach, incorporating 32 different computational paths that include feature column rotation, class label rotation, and power transformations to enhance prediction robustness and accuracy.
    \item \textbf{LightGBM and Other Methods:} These models are well-regarded for their efficiency in handling tabular data and have been included for benchmarking purposes.
\end{itemize}

\subsection{Implementation details}
We did not to re-test TabPFN or other models such as XGBoost and AdaBoost. Instead, we used the results from \cite{hollmann2023tabpfn} for comparison. Each of our models was tested using a consistent methodology as described in \cite{hollmann2023tabpfn} to ensure a fair comparison across all tests. For each dataset, we evaluated 5 repetitions, 50\% train and 50\% test samples, each with a different random seed and train-test split. Given the same seeds, our test employs the same split as in \cite{hollmann2023tabpfn}.

 We are careful to prevent data leakage. For each split of each dataset, our model is fine-tuning on the train set and evaluated on the test set using the model which performs best on the train set. Importantly, the test set does not play a role in selecting the model.We follow the cross-entropy loss function used in the original TabPFN.

\subsection{Results}

Our results are in Table \ref{my_tabular_results_table}, which includes comparisons with other models whose data were derived from prior studies \cite{hollmann2023tabpfn}. Using AUC ROC OVO as the evaluation metric (a standard choice for multi-class classification problems), the FT-TabPFN model we developed consistently outperformed other models in three of the five datasets tested. Notably, its average AUC ROC OVO ranking is tied with XGBoost for first place, both at 2.6, highlighting its effectiveness.

\textbf{Performance Highlights:}
\begin{itemize}
    \item Across all datasets, every version of the FT-TabPFN model outperforms the baseline TabPFN$_{n.e.}$, demonstrating the overall effectiveness of the feature tokenization approach.
    
    \item The full version of FT-TabPFN was the top performer across all datasets, confirming its superior ability to handle complex categorical structures. This model incorporates both feature identifiers and orthogonal regularisation, highlighting the essential role of these enhancements in improving performance.
    
    \item Comparative analysis between versions of FT-TabPFN showed incremental benefits: The model with feature identifiers (FT-TabPFN${n.r.}$), although lacking orthogonal regularisation, still outperformed the version without identifiers (FT-TabPFN${n.i.}$). This underlines the significant impact of feature identifiers on model performance. The full version, including both feature identifiers and orthogonal regularisation, outperformed all other configurations. This suggests that while feature identifiers alone provide a significant performance boost, the addition of orthogonal regularisation maximises the potential of the model by improving feature discrimination.
\end{itemize}

\begin{figure*}
  \centering
  \begin{subfigure}[b]{0.3\linewidth}
    \centering
    \includegraphics[width=\linewidth]{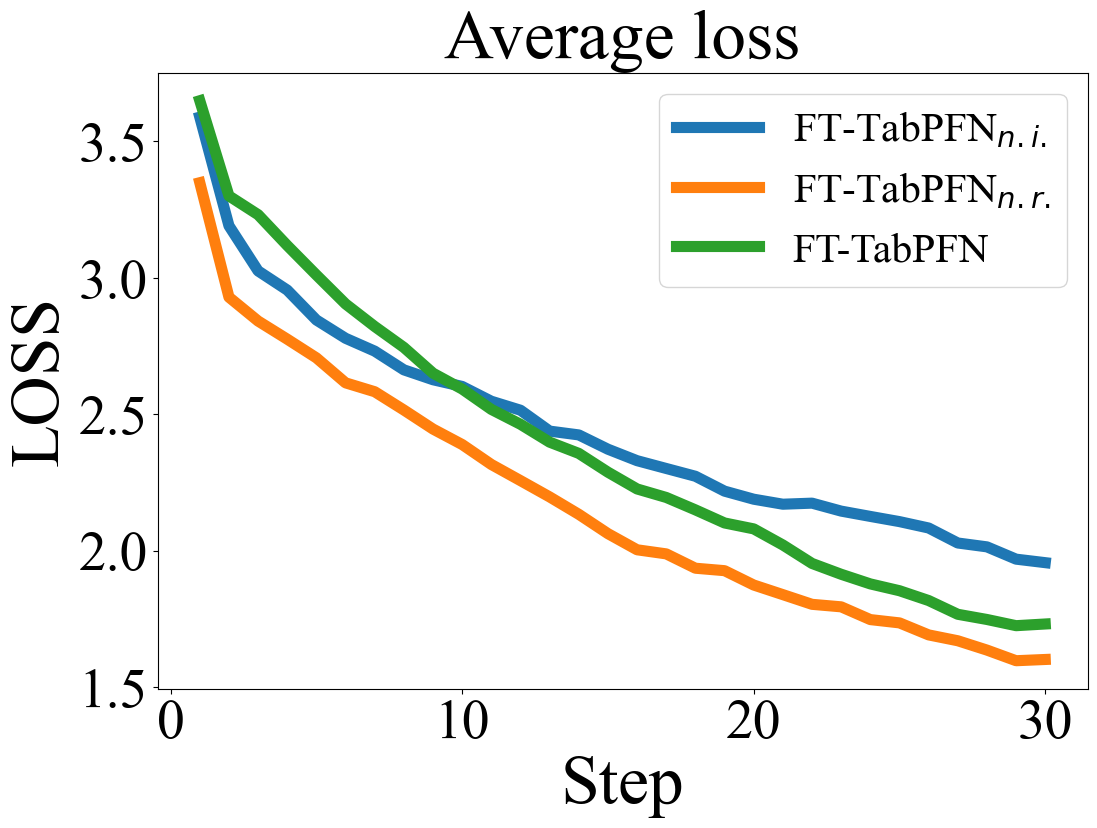}
    \caption{Average Train Loss Across All Datasets}
  \end{subfigure}%
  \hfill
  \begin{subfigure}[b]{0.3\linewidth}
    \centering
    \includegraphics[width=\linewidth]{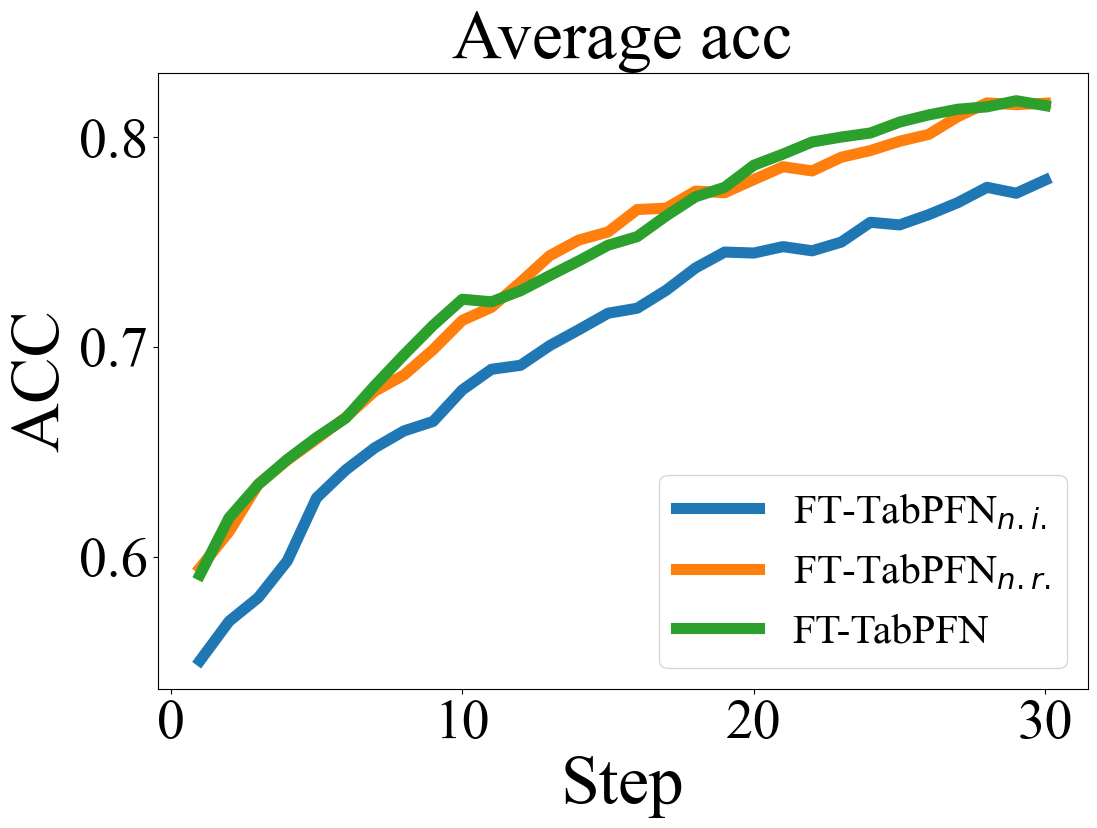}
    \caption{Average Train ACC Across All Datasets}
  \end{subfigure}%
  \hfill
  \begin{subfigure}[b]{0.3\linewidth}
    \centering
    \includegraphics[width=\linewidth]{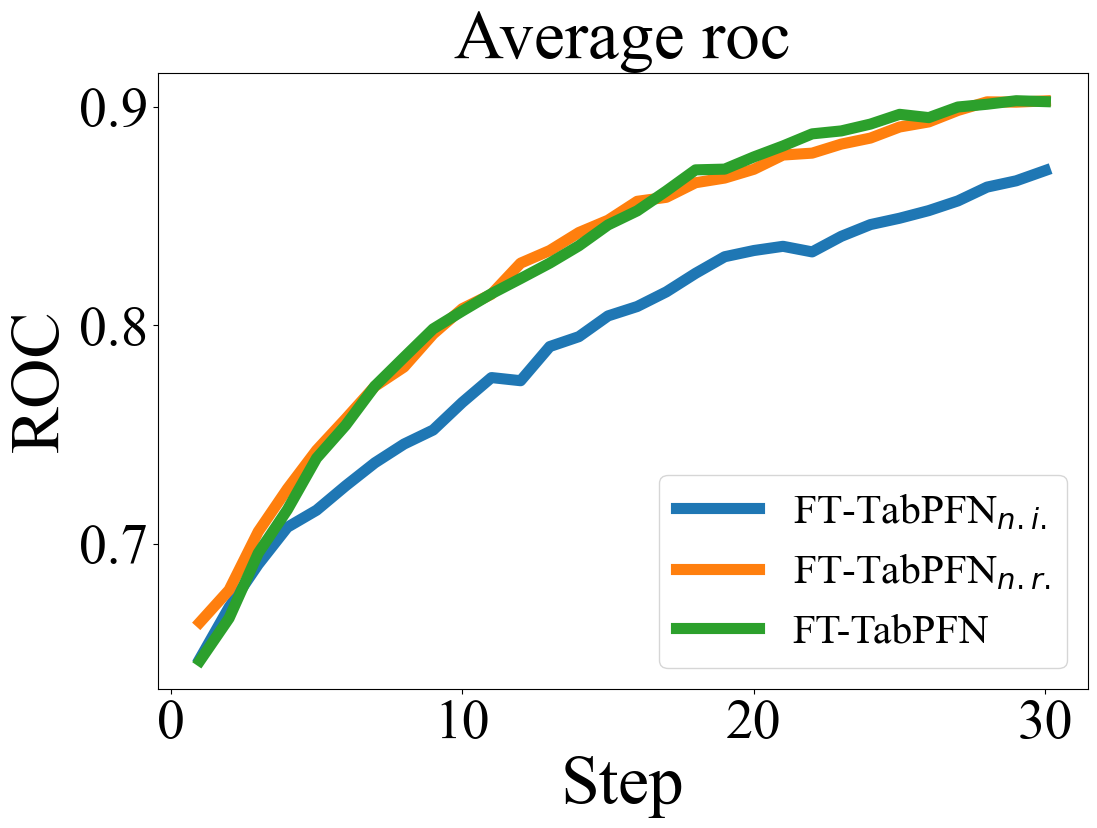}
    \caption{Average Train ROC Across All Datasets}
  \end{subfigure}%
  \\
  \vspace{1em} 
  \begin{subfigure}[b]{0.3\linewidth}
    \centering
    \includegraphics[width=\linewidth]{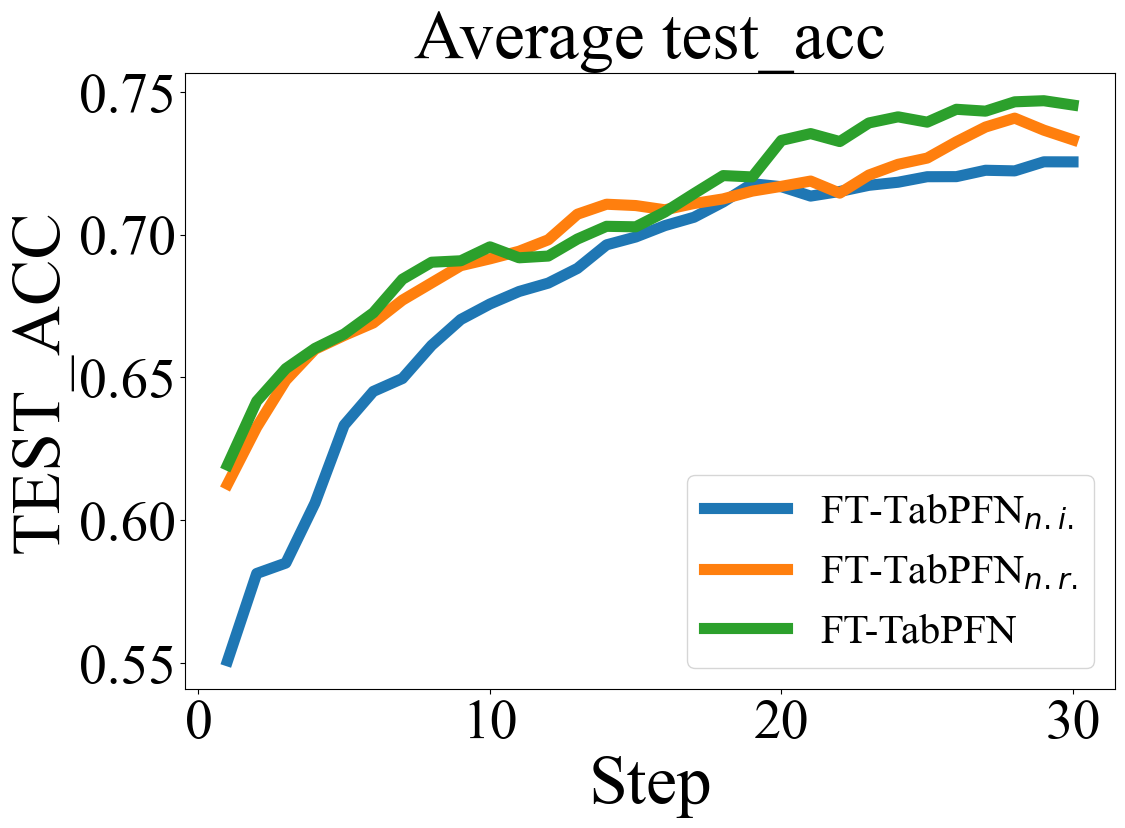}
    \caption{Average Test ACC Across All Datasets}
  \end{subfigure}%
  \hspace{0.15\linewidth} 
  \begin{subfigure}[b]{0.3\linewidth}
    \centering
    \includegraphics[width=\linewidth]{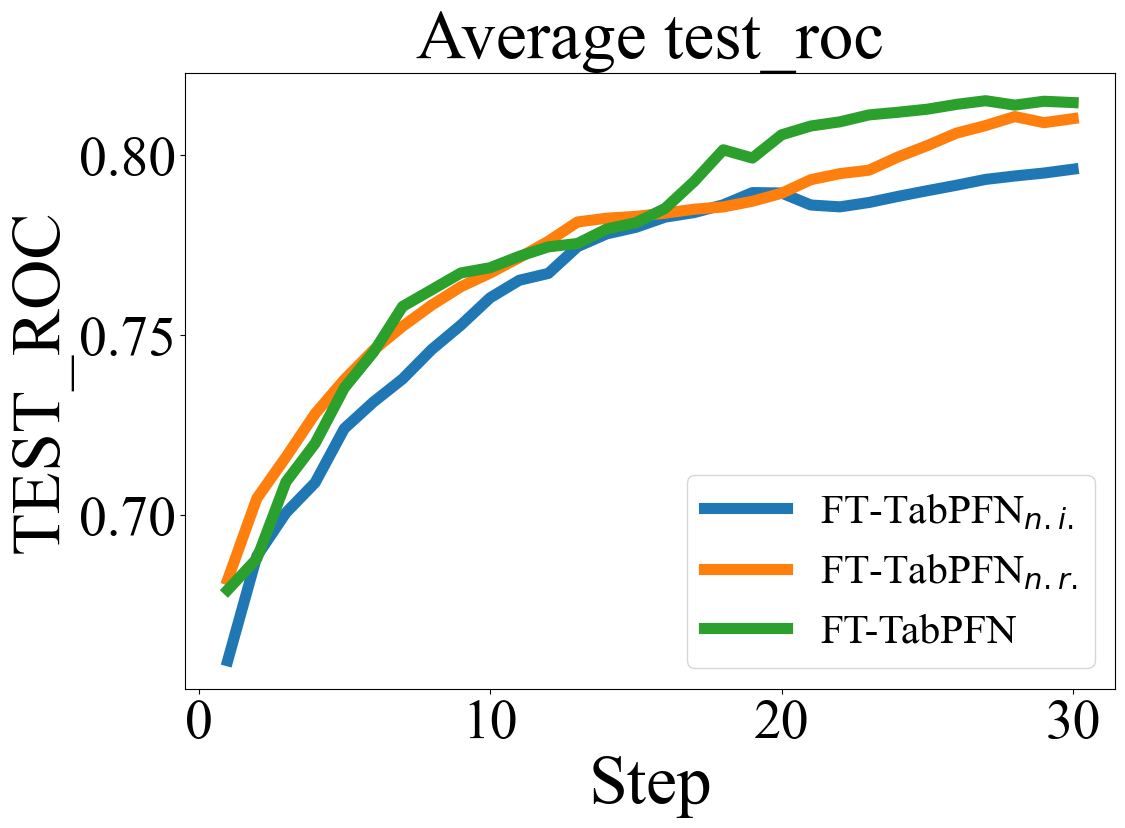}
    \caption{Average Test ROC Across All Datasets}
  \end{subfigure}
  \caption{\textbf{Average training and test logs over all datasets.} These graphs show the average training and testing metrics for all datasets, comparing FT-TabPFN$_{n.i.}$, FT-TabPFN$_{n.r.}$, and FT-TabPFN.}
  \label{fig:trainlog}
\end{figure*}

\begin{figure*}
  \centering
  \begin{subfigure}[b]{1\linewidth}
    \centering
    \includegraphics[width=\linewidth]{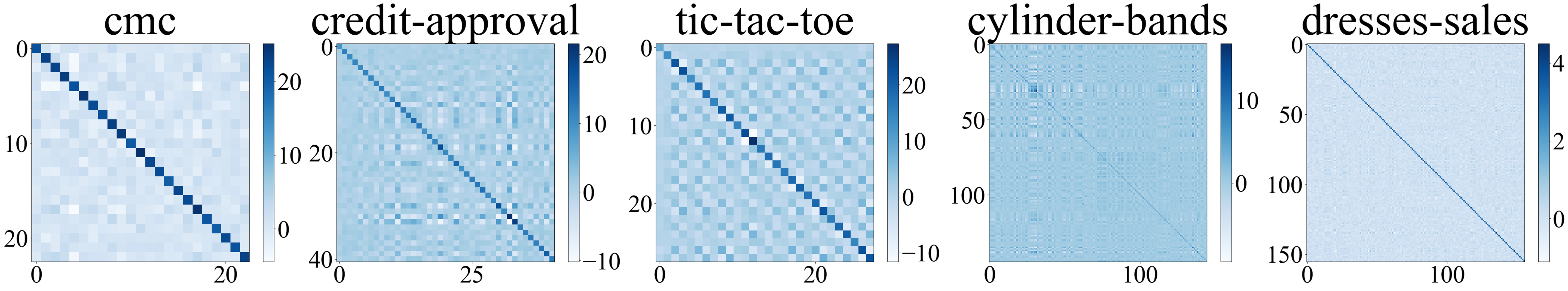}
    \caption{Without feature identifiers}
    \label{fig:correlationmap:no_IR}
  \end{subfigure}
  \begin{subfigure}[b]{1\linewidth}
    \centering
    \includegraphics[width=\linewidth]{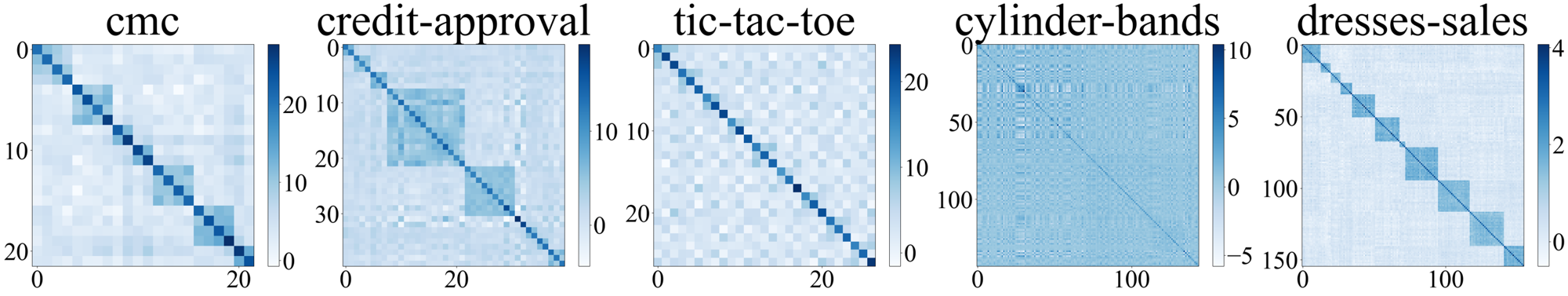}
    \caption{With feature identifiers, without regularization}
    \label{fig:correlationmap:I}
  \end{subfigure}
  \begin{subfigure}[b]{1\linewidth}
    \centering
    \includegraphics[width=\linewidth]{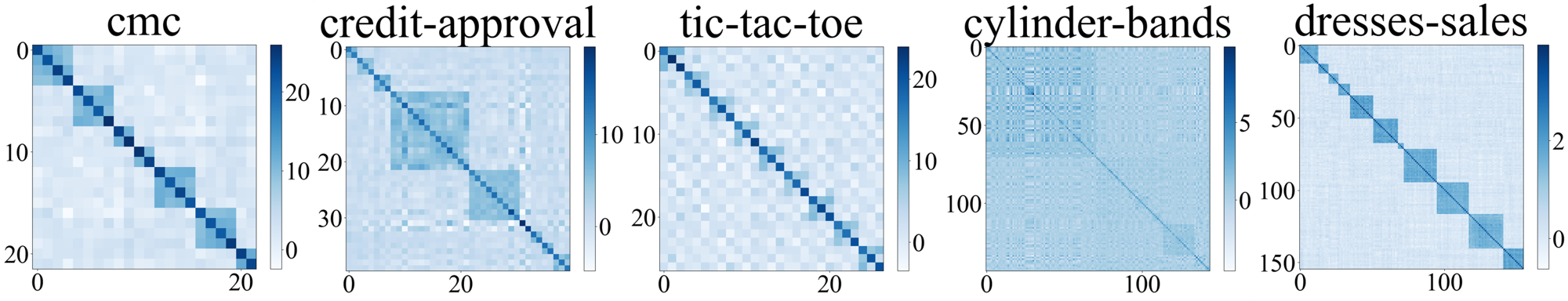}
    \caption{With feature identifiers and regularization}
    \label{fig:correlationmap:IR}
  \end{subfigure}
  \caption{\textbf{Category Tokens Correlation Heatmaps.}
  Each subgraph is the inner product matrix of all category tokens under the classification characteristics of each data set. These heatmaps illustrate the inner product matrices of category tokens under different configurations, showing changes in category tokens similarities.}
  \label{fig:correlationmap}
\end{figure*}

\begin{figure*}
  \centering
  \begin{subfigure}[b]{1\linewidth}
    \centering
    \includegraphics[width=\linewidth]{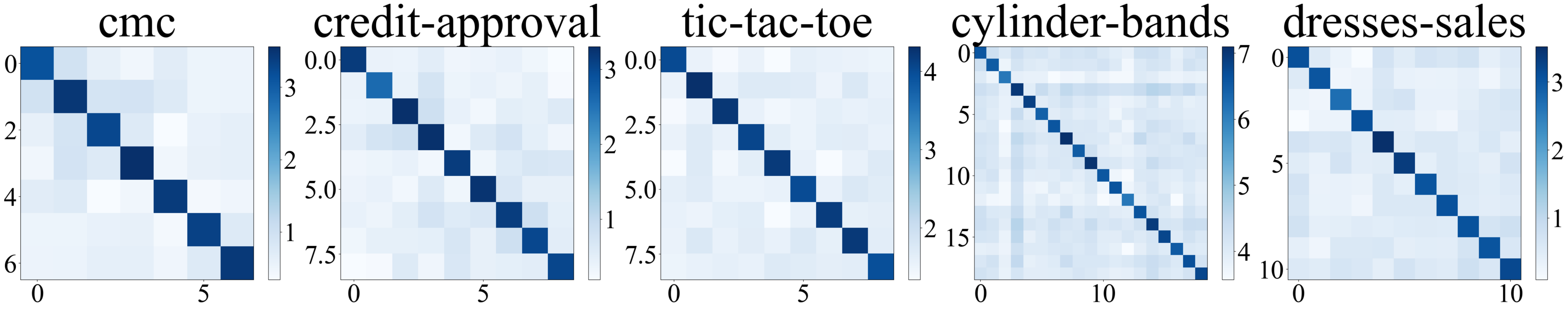}
    \caption{Without orthogonal regularization}
    \label{fig:FIcorrelationmap:I}
  \end{subfigure}
  \begin{subfigure}[b]{1\linewidth}
    \centering
    \includegraphics[width=\linewidth]{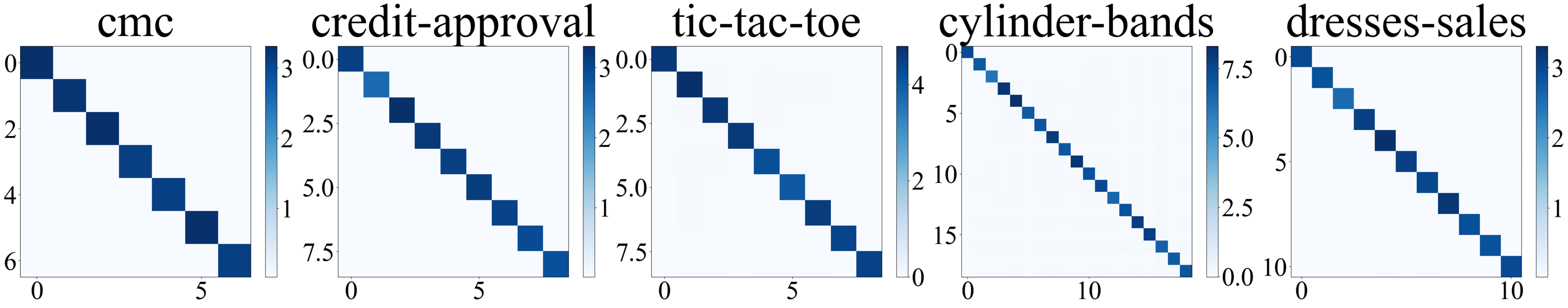}
    \caption{With orthogonal regularization}
    \label{fig:FIcorrelationmap:IR}
  \end{subfigure}
  \caption{\textbf{Feature Identifier Correlation Heatmaps.} These maps show the changes in correlation between different feature identifiers with and without orthogonal regularization.}
  \label{fig:FIcorrelationmap}
\end{figure*}

\section{analysis}
\label{sec:analysis}
The focus of our analysis is to elucidate the differences in performance between the three versions of the FT-TabPFN model, specifically examining the role of feature identifiers and orthogonal regularization. The detailed performance comparison can be observed in Fig.\ref{fig:trainlog}, where complete FT-TabPFN model (green curve) shows superior performance metrics during both training and testing phases compared to the other two models (FT-TabPFN${n.i.}$ is the blue curve and FT-TabPFN${n.r.}$ is the yellow curve).

\textbf{Understanding Feature Identifiers without Regularization:}
Feature identifiers significantly enhance model differentiation capabilities by embedding commonalities within features while distinguishing between them. This mechanism is clearly illustrated in Fig.\ref{fig:correlationmap:no_IR} and Fig.\ref{fig:correlationmap:I}, where the introduction of feature identifiers significantly increases the similarity of categories within features (consecutive indices).

\textbf{Role of Orthogonal Regularization:}
Orthogonal regularization refines the separation induced by feature identifiers by enforcing a more pronounced orthogonality between them. This is evident in the comparison between Fig.\ref{fig:FIcorrelationmap:I} (without orthogonal regularization) and Fig.\ref{fig:FIcorrelationmap:IR} (with orthogonal regularization), where the latter displays significantly lower cross-identifier correlations.

\textbf{Comprehensive Analysis:}
The comprehensive category token analysis with orthogonal regularization (Fig.\ref{fig:correlationmap:IR}) shows more defined patterns of correlation within features and reduced correlation across features, aligning with the orthogonal regularization's objectives.

The structured heatmaps (Figs. \ref{fig:correlationmap} and \ref{fig:FIcorrelationmap}) visually underscore the theoretical improvements posited by the introduction of feature identifiers and their regularization, providing compelling evidence of their beneficial impact on the FT-TabPFN model's performa

\section{conclusion}

This study presents FT-TabPFN, an improved version of the TabPFN model that is specifically designed to address the challenges of classifying tabular data with categorical features. The FT-TabPFN model includes a novel feature tokenization layer that handles numerical and categorical features differently, enhancing the model's ability to handle the inherent diversity of tabular data.

The experiments conducted on various datasets with categorical features demonstrate that FT-TabPFN outperforms the original TabPFN model significantly. The incorporation of feature identifiers and their orthogonal regularization further enhances the performance and enables the model to distinguish between different categories more efficiently. This is crucial for improving the accuracy of the classification task.

In conclusion, this paper proposes the FT-TabPFN model, which provides a more nuanced approach to feature representation in tabular data. This approach can better exploit the complexity of tabular datasets. Future work could consider optimizing the structure and parameters of the feature tokenization layer further. Additionally, exploring different kinds of orthogonal regularization methods and extending the model to accommodate larger datasets and more diverse application scenarios could be beneficial.

\bibliographystyle{unsrt}  
\bibliography{references}

\end{document}